# Support Vector Machine Classifier with Rescaled Huberized Pinball Loss

Shibo Diao, Department of Mathematics, Beijing Jiaotong University, 100044 Beijing, China,

**ABSTRACT**

Support vector machines are widely used in machine learning classification tasks, but traditional SVM models suffer from sensitivity to outliers and instability in resampling, which limits their performance in practical applications. To address these issues, this paper proposes a novel rescaled Huberized pinball loss function with asymmetric, non-convex, and smooth properties.

Based on this loss function, we develop a corresponding SVM model called RHPSVM (Rescaled Huberized Pinball Loss Support Vector Machine). Theoretical analyses demonstrate that RHPSVM conforms to Bayesian rules, has a strict generalization error bound, a bounded influence function, and controllable optimality conditions, ensuring excellent classification accuracy, outlier insensitivity, and resampling stability.

Additionally, RHPSVM can be extended to various advanced SVM variants by adjusting parameters, enhancing its flexibility. We transform the non-convex optimization problem of RHPSVM into a series of convex subproblems using the concave-convex procedure (CCCP) and solve it with the ClipDCD algorithm, which is proven to be convergent. Experimental results on simulated data, UCI datasets, and small-sample crop leaf image classification tasks show that RHPSVM outperforms existing SVM models in both noisy and noise-free scenarios, especially in handling high-dimensional small-sample data.

**INTRODUCTION**

**1.1 Research Background**

In the era of big data, machine learning has become an essential tool for data processing and pattern recognition. Support vector machines (SVMs), proposed by Vapnik and Cortes[1], are widely used in image recognition[3], biomedicine[4], industrial defect detection[5], and financial risk prediction[6] due to their superior generalization performance based on statistical learning theory[2]. However, traditional SVMs based on the hinge loss function have two critical limitations:

1. Outlier sensitivity: The hinge loss is unbounded, making the model vulnerable to interference from outliers and label noise.

2. Resampling instability: The decision boundary of SVMs depends on support vectors, leading to significant fluctuations in model performance under resampling.

To overcome these drawbacks, researchers have proposed various improved loss functions. For example, Huang[8] introduced the pinball loss to enhance resampling stability, and Xu[9] proposed the rescaled hinge loss to improve outlier robustness. However, existing loss functions still face challenges such as non-smoothness, lack of sparsity, or limited flexibility.

**1.2 Research Contributions**

This paper makes the following key contributions:

1. A novel rescaled Huberized pinball loss function is proposed, which integrates the advantages of Huberized pinball loss and correntropy, achieving asymmetry, non-convexity, and smoothness.

2. The RHPSVM model is established, which can be extended to multiple advanced SVM variants (e.g., PINSVM, RQSVM) by adjusting parameters, enhancing adaptability to different data scenarios.

3. Theoretical proofs verify that RHPSVM has Bayesian consistency, a strict generalization error bound, bounded influence function, and resampling stability, ensuring its robustness and reliability.

4. A CCCP-based solution algorithm with proven convergence is designed to solve the non-convex optimization problem of RHPSVM.

## 1.3 Paper Structure

The remainder of this paper is organized as follows: Section 2 reviews related theories of SVM and loss functions. Section 3 details the construction of the rescaled Huberized pinball loss and the RHPSVM model, along with theoretical analyses. Section 4 concludes the paper.

## TRADITIONAL METHODS

### 2.1 Support Vector Machine Fundamentals

SVMs aim to find the optimal separating hyperplane that maximizes the margin between two classes[2]. Linear SVMs are suitable for linearly separable data, while kernel tricks[7] enable SVMs to handle non-linear problems by mapping data to high-dimensional feature spaces. Soft margin SVMs[1] introduce slack variables to tolerate minor misclassifications, using loss functions to measure classification errors.

### 2.2 Improved Loss Functions for SVM

In 1995, Cortes and Vapnik used the hinge loss function[1] to establish the first soft-margin SVM model, which was called the hinge support vector machine (HSVM), and its form is as follows:

$$\min_{w,b} \frac{1}{2}||w||_2^2 + C \sum_{i=1}^{n} L_{\text{hinge}}(1 - y_i(w^T x_i + b)) \quad (1)$$

where $C > 0$ represents the penalty parameter, which is used to measure the model's tolerance for classification errors. HSVM employs the hinge loss function.

$$L_{hinge}(u) = \begin{cases} u & u \geq 0 \\ 0 & u < 0 \end{cases} \quad (2)$$

Introducing slack variables to (1) can transform it into

$$\min_{w,b} \frac{1}{2}|w|_2^2 + C \sum_{i=1}^{n} \xi_i$$

$$s.t. y_i(w^T x_i + b) \geq 1 - \xi_i, \quad (3)$$

$$\xi_i \geq 0, i = 1 \dots n$$

HSVM attempts to maximize the shortest distance between the two classes. Therefore, the decision hyperplane $w^T x_i + b = \pm 1$ is determined only by a small number of support vectors, so the resulting classifier is relatively sensitive to boundary noise.

Pinball loss is proposed by Huang[8], it replaces the shortest distance with quantile distance to enhance resampling stability but suffers from singularity at zero. The form of pinball loss is

$$L_\tau(u) = \begin{cases} u & u \geq 0 \\ -\tau u & u < 0 \end{cases} \quad (4)$$

where $\tau \in [0,1]$ represents the quantile parameter. When $\tau = 1$ the pinball loss becomes $l_1$ loss, so the pinball loss is also the generalized $l_1$ loss. The form of PINSVM is

$$\min_{w,b} \frac{1}{2}||w||_2^2 + C \sum_{i=1}^{n} L_\tau(1 - y_i(w^T x_i + b)) \quad (5)$$

Similarly, (5) can be equivalently transformed into

$$\min_{w,b} \frac{1}{2}|w|_2^2 + C \sum_{i=1}^{n} \xi_i$$

$$s.t. y_i(w^T x_i + b) \geq 1 - \xi_i, \quad (6)$$

$$y_i(w^T x_i + b) \leq 1 + \frac{1}{\tau}\xi_i, i = 1 \dots n$$

Zhu[12] proposed Huberized pinball loss, which improved the pinball loss to achieve smoothness, but it lacks outlier robustness. The form of Huberized pinball loss is

$$L_{rhp}(u) = \begin{cases} u - \frac{s}{2} & u > s \\ \frac{u^2}{2s} & 0 < u \leq s \\ \frac{\tau u^2}{2s} & -s < u \leq 0 \\ \tau\left(-u - \frac{s}{2}\right) & u \leq -s \end{cases} \quad (7)$$

Where $s > 0$ is a preset constant and $\tau \in [0,1]$ is the quantile parameter. When appropriate parameters are set, the Huber ball loss can be reduced to the ball loss and the $l_2$ loss. It is also worth noting that, unlike the ball loss, the Huber ball loss is differentiable everywhere.

Xu[9] proposed rescaled hinge loss and generalized quantile loss based on correntropy, improving outlier resistance but with limited flexibility. However, existing loss functions cannot simultaneously achieve outlier insensitivity, resampling stability, smoothness, and flexibility. This paper addresses this gap by proposing the rescaled Huberized pinball loss.

**RHPSVM MODEL AND THEORETICAL ANALYSIS**

## 3.1 Rescaled Huberized Pinball Loss Function

Combining the Huberized pinball loss[12] and correntropy theory[13], we construct the Rescaled Huberized Pinball loss function as follows:

$$L_{rhp}(u) = \begin{cases} \eta(1 - \exp(-(u - s/2)/\lambda)) & u > s \\ \eta(1 - \exp(-u^2/2s\lambda)) & 0 < u \leq s \\ \eta(1 - \exp(-\tau u^2/2s\lambda)) & -s < u \leq 0 \\ \eta(1 - \exp((u + s/2)/\lambda)) & u \leq -s \end{cases} \quad (8)$$

Figure 1.

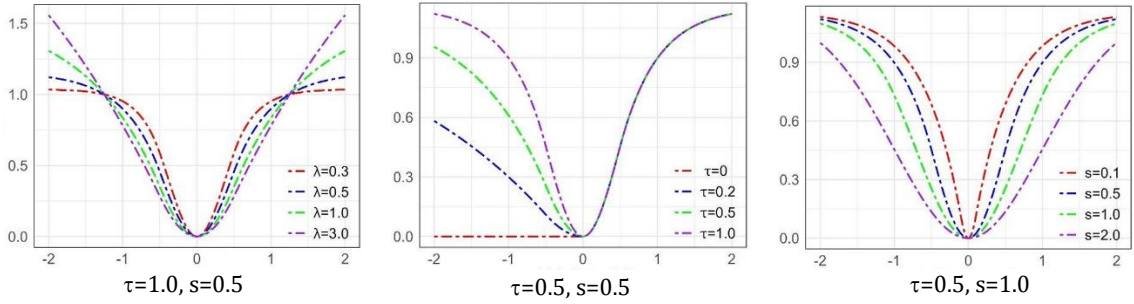

τ=1.0, s=0.5     τ=0.5, s=0.5     τ=0.5, s=1.0

Figure 1 show the influence of the change in the parameters, τ and s, on $L_{rhp}(u)$. From the expression and graph of $L_{rhp}(u)$, it can be seen that the re-weighted Huber ball loss is asymmetric, non-convex and smooth.

### 3.2 RHPSVM Model Formulation

Under the soft margin SVM regularization framework, RHPSVM is defined as:

$$\min_{w,b} \frac{1}{2} w^T w + C \sum_{i=1}^{n} L_{rhp}(1 - y_i w^T \phi(x_i)), \quad (9)$$

where $\tilde{w} = (w^T, b)^T$, $\tilde{x} = (x^T, e)$, C>0 is the penalty parameter $\phi(\tilde{x}) = (\phi(x)^T, 1)$ is the kernel mapping.

### 3.3 Theoretical Properties

#### 3.3.1 Bayesian Consistency

Theorem 3.1: For $\lambda > 0, s > 0$ if $0 \leq \tau \leq 1$, $L_{rhp}(u)$ is Fisher consistent, and the optimal classifier $f^*(x)$ derived from minimizing the expected risk $E[L_{rhp}(1 - yf(x))]$ is the Bayesian classifier.

Proof: First, according to (8), suppose $u = 1 - z$, and $L_{rhp}(z) = L_{rhp}(1 - z)$, then we will get

$$L_{rhp}(z) = \begin{cases} \eta\left(1 - \exp\left(\frac{\tau\left(1 - z + \frac{s}{2}\right)}{\lambda}\right)\right) & z > 1 + s \\ \eta\left(1 - \exp\left(-\frac{\tau(1-z)^2}{2s\lambda}\right)\right) & 1 < z \leq 1 + s \\ \eta\left(1 - \exp\left(-\frac{(1-z)^2}{2s\lambda}\right)\right) & 1 - s < z \leq 1 \\ \eta\left(1 - \exp\left(-\frac{1 - z - \frac{s}{2}}{\lambda}\right)\right) & z \leq 1 - s \end{cases} \quad (10)$$

Next, we will discuss the relationship between $L_{rhp}(z)$ and $L_{rhp}(-z)$ in three different scenarios, where $z > 0$.

(1) when $z > 1 + s$, that is $-z < -1 - s$, we will get

$$L_{rhp}(z) = \eta\left(1 - \exp\left(\frac{\tau\left(1-z+\frac{s}{2}\right)}{\lambda}\right)\right)$$

$$L_{rhp}(-z) = \eta\left(1 - \exp\left(\frac{\tau\left(1+z-\frac{s}{2}\right)}{\lambda}\right)\right)$$

When $0 \leq \tau \leq 1$, $\tau\left(1-z+\frac{s}{2}\right) < \left(1+z-\frac{s}{2}\right)$.
And also because $\left[1 - \exp\left(\frac{u}{\lambda}\right)\right]$ is monotonically increasing on u, $L_{rhp}(z) < L_{rhp}(-z)$

(2) when $1 < z \leq 1+s$
that is $-1-s \leq -z < -1$, we will get

$$L_{rhp}(z) = \eta\left(1 - \exp\left(-\frac{\tau(1-z)^2}{2s\lambda}\right)\right)$$

For $L_{rhp}(-z)$, we should discuss the range of s. When $1-s > -1$,

$$L_{rhp}(-z) = \eta\left(1 - \exp\left(-\frac{\tau\left(1+z-\frac{s}{2}\right)}{\lambda}\right)\right),$$

we can easily get $\frac{\tau(1-z)^2}{2s} < 1+z-\frac{s}{2}$,
then $L_{rhp}(z) < L_{rhp}(-z)$.
When $1-s < -1$,

$L_{rhp}(z)$
$$= \begin{cases} \eta\left(1 - \exp\left(-\frac{(1+z)^2}{2s\lambda}\right)\right) & 1-s < -z \leq -1 \\ \eta\left(1 - \exp\left(-\frac{\left(1+z-\frac{s}{2}\right)}{\lambda}\right)\right) & -1-s < -z \leq 1-s \end{cases}$$

We can easily get $\tau(1-z)^2 < (1+z)^2$. And when $-1-s < -z \leq 1-s$, we can also get $\frac{\tau(1-z)^2}{2s} < 1+z-\frac{s}{2}$, then $L_{rhp}(z) < L_{rhp}(-z)$. So, when $1 < z \leq 1+s$, we have $L_{rhp}(z) < L_{rhp}(-z)$.

(3) when $0 < z \leq 1$ that is $-1 \leq -z < 0$, when $1-s < 0$, $L_{rhp}(z) = \eta\left(1 - \exp\left(-\frac{(1-z)^2}{2s\lambda}\right)\right)$. For $L_{rhp}(-z)$, we should discuss the range of s. When $1-s < -1$, $L_{rhp}(-z) = \eta\left(1 - \exp\left(-\frac{(1+z)^2}{2s\lambda}\right)\right)$,
we can easily get $\frac{(1-z)^2}{2s} < \frac{(1+z)^2}{2s}$,
when $1-s > -1$,

$L_{rhp}(-z)$
$$= \begin{cases} \eta\left(1 - \exp\left(-\frac{(1+z)^2}{2s\lambda}\right)\right) & 1-s < -z \leq 0 \\ \eta\left(1 - \exp\left(-\frac{\left(1+z-\frac{s}{2}\right)}{\lambda}\right)\right) & -1 < -z \leq 1-s \end{cases}$$

we can easily get $\frac{(1-z)^2}{2s} < \frac{(1+z)^2}{2s}$.
when $-1 \leq -z < 1-s$, we can also get $\frac{(1-z)^2}{2s} < 1+z-\frac{s}{2}$, so when $s > 1$, we get $L_{rhp}(z) < L_{rhp}(-z)$.
when $1-s > 0$,

$L_{rhp}(z)$
$$= \begin{cases} \eta\left(1 - \exp\left(-\frac{(1-z)^2}{2s\lambda}\right)\right) & 1-s < z \leq 1 \\ \eta\left(1 - \exp\left(-\frac{\left(1-z-\frac{s}{2}\right)}{\lambda}\right)\right) & 0 < z \leq 1-s \end{cases}$$

$L_{rhp}(-z) = \eta\left(1 - \exp\left(\frac{-\left(1+z-\frac{s}{2}\right)}{\lambda}\right)\right)$, which is obviously true for $\left(1-z-\frac{s}{2}\right) < \left(1+z-\frac{s}{2}\right)$.

When $1-s < z \leq 1$, we can also get $\frac{(1-z)^2}{2s} < 1+z-\frac{s}{2}$, so when $0 < s < 1$, we have $L_{rhp}(z) < L_{rhp}(-z)$.

on the other hand, the first-order derivative of $L_{rhp}(z)$ with respect to z is

$L'_{rhp}(z)$
$$= \begin{cases} \frac{\eta\tau}{\lambda}\left(1 - \exp\left(\frac{\tau\left(1-z+\frac{s}{2}\right)}{\lambda}\right)\right) & z > 1+s \\ -\frac{\tau\eta(1-z)}{\lambda s}\exp\left(-\frac{\tau(1-z)^2}{2s\lambda}\right) & 1 < z \leq 1+s \\ -\frac{\tau\eta(1-z)}{\lambda s}\exp\left(-\frac{(1-z)^2}{2s\lambda}\right) & 1-s < z \leq 1 \\ -\frac{\eta}{\lambda}\exp\left(-\frac{1-z-\frac{s}{2}}{\lambda}\right) & z \leq 1-s \end{cases}$$

We can easily get $L'_{rhp}(0) \neq 0$. By verifying the conditions of Lin's theorem[14], we prove that $L_{rhp}(z) < L_{rhp}(-z)$ for all $z > 0$ and $L'_{rhp}(0) \neq 0$ satisfying Fisher consistency. So $f^*(x)$ is the Bayesian classifier.

### 3.3.2 Generalization Error Bound

Theorem 3.2: With probability at least $1-\zeta$, RHPSVM satisfies:

$$P(yf(x) \le 0) \le \frac{\gamma}{n}\sum_{i=1}^{n} L_{rhp}(1 - y_i f(x_i))$$

$$+ \frac{2B\gamma\iota}{\sqrt{n}}\sqrt{\sum_{i=1}^{n} K(x_i, x_i)} + \sqrt{\frac{8\ln\left(\frac{2}{\zeta}\right)}{n}} \quad (11)$$

where $\iota$ is the Lipschitz constant of $L_{rhp}$, and B is the upper bound of $||\widetilde{w}||$.

Proof: Derived using Rademacher complexity[15] and Lipschitz continuity of $L_{rhp}$.

### 3.4 Solution Algorithm

The non-convex optimization problem of RHPSVM is transformed into convex subproblems using CCCP [16]. The ClipDCD algorithm[10] is adopted for efficient solving, with convergence proven by the monotonic decreasing property of the objective function.

By observing the expression of $L_{rhp}(u)$ for loss can be decomposed into the following two functions:

$$g(u) = \frac{\eta}{\lambda} L_{hp}(u), h(u)$$
$$= -\frac{\eta}{\lambda} L_{hp}(u) + \eta\left(1 - exp\left(-\frac{L_{hp}(u)}{\lambda}\right)\right) \quad (12)$$

Obviously, g(u) is a convex function and h(u) is a concave function here. By substituting them into (9), the problem can be transformed into

$$\min_{w,b} \frac{1}{2}\widetilde{w}^T\widetilde{w} + C\sum_{i=1}^{n} g\left(1 - y_i\widetilde{w}^T\phi(\widetilde{x}_i)\right)$$
$$+ C\sum_{i=1}^{n} h\left(1 - y_i\widetilde{w}^T\phi(\widetilde{x}_i)\right) \quad (13)$$

Furthermore, we can use the concave convex process to optimize problem (13) by transforming the original RHPSVM problem into a series of convex subproblems for solution.

$$\widetilde{w}^{k+1} = arg\min_{\widetilde{w}} L_{vex}(\widetilde{w}) + \nabla L_{cav}(\widetilde{w}^k)^T \cdot \widetilde{w} \quad (14)$$

where $L_{vex}$ and $L_{cav}$ correspond to the convex and concave parts in (13), respectively. $\nabla L_{cav}(\widetilde{w}^k)$ represents the derivative of $\nabla L_{cav}(\widetilde{w})$ with respect to the optimal solution $\widetilde{w}^k$. To simplify the process, we introduce an auxiliary variable $\delta^k = (\delta_1^k, \dots, \delta_n^k)^T$ as follows:

$$\delta_i^k = \begin{cases} \frac{\eta}{\lambda}\left(1 - \exp\left(-\frac{1 - y_i\widetilde{w}^T\phi(\widetilde{x}_i) - \frac{s}{2}}{\lambda}\right)\right) & 1 - y_i\widetilde{w}^T\phi(\widetilde{x}_i) \ge s \\ \frac{\eta(1 - y_i\widetilde{w}^T\phi(\widetilde{x}_i))}{\lambda s}\left(1 - exp\left(-\frac{(1 - y_i\widetilde{w}^T\phi(\widetilde{x}_i))^2}{2s\lambda}\right)\right) & 0 < 1 - y_i\widetilde{w}^T\phi(\widetilde{x}_i) < s \\ \frac{\tau\eta(1 - y_i\widetilde{w}^T\phi(\widetilde{x}_i))}{\lambda s}\left(1 - exp\left(-\frac{(1 - y_i\widetilde{w}^T\phi(\widetilde{x}_i))^2}{2s\lambda}\right)\right) & -s < 1 - y_i\widetilde{w}^T\phi(\widetilde{x}_i) < 0 \\ -\frac{\tau\eta}{\lambda}\left(1 - \exp\left(\frac{\tau(1 - y_i\widetilde{w}^T\phi(\widetilde{x}_i) + \frac{s}{2})}{\lambda}\right)\right) & 1 - y_i\widetilde{w}^T\phi(\widetilde{x}_i) \le -s \end{cases} \quad (15)$$

then $\nabla h(1 - y_i\widetilde{w}^T\phi(\widetilde{x}_i)) = \delta_i^k \cdot y_i\phi(x_i)$, So applying the concave convex process can transform problem (13) into

$$\min_{w,b} \frac{1}{2}\widetilde{w}^T\widetilde{w} + C\sum_{i=1}^{n} g\left(1 - y_i\widetilde{w}^T\phi(\widetilde{x}_i)\right)$$
$$+ C\sum_{i=1}^{n} \delta_i^k \cdot y_i\phi^T(\widetilde{x}_i)\widetilde{w} \quad (16)$$

After introducing slack variables, the problem becomes

$$\min_{w,b} \frac{1}{2}\widetilde{w}^T\widetilde{w} + C\sum_{i\in I_1}\xi_i^2 + C\sum_{i\in I_2}\xi_i$$
$$+ C\sum_{i=1}^{n}\delta_i^k \cdot y_i\phi^T(\widetilde{x}_i)\widetilde{w}$$

$$s.t.\ 1 - y_i\widetilde{w}^T\phi(\widetilde{x}_i) \le \frac{2s\lambda}{\eta}\xi_i$$

$$1 - y_i\widetilde{w}^T\phi(\widetilde{x}_i) \le \frac{s}{2} + \frac{\lambda}{\eta}\xi_i, \quad (17)$$

$$-1 + y_i\widetilde{w}^T\phi(\widetilde{x}_i) \le \frac{2s\lambda}{\tau\eta}\xi_i,$$

$$-1 + y_i\widetilde{w}^T\phi(\widetilde{x}_i) \le \frac{s}{2} + \frac{\lambda}{\tau\eta}\xi_i, i \in I_1, j \in I_2.$$

where $I_1 = \{i||1 - y_i\widetilde{w}^T\phi(\widetilde{x}_i)| < s\}$,

$I_2 = \{j||1 - y_i\widetilde{w}^T\phi(\widetilde{x}_i)| \ge s\}$.

Further, (17) can be transformed into matrix form

$$\min_{w,b} \frac{1}{2}\widetilde{w}^T\widetilde{w} + C\xi^T E_1\xi + CeE_2\xi + C\widetilde{w}^T Y A^T \delta^k$$

$$s.t. e - YA^T\widetilde{w} \leq \frac{\lambda}{\eta}(2sE_1 + E_2)\xi + \frac{s}{2}E_2 e, \quad (18)$$

$$-e + YA^T\widetilde{w} \leq \frac{\lambda}{\eta\tau}(2sE_1 + E_2)\xi + \frac{s}{2}E_2 e,$$

where e is an n-dimensional unit vector,

$Y = \text{diag}(y_1, \ldots, y_n)$ and $A = (\phi(\widetilde{x_1}), \ldots \phi(\widetilde{x_n}))$. And $E_1, E_2$ is a diagonal matrix that when $i = j \in I_1, (E_1)_{ij} = 1$, otherwise $(E_1)_{ij} = 0$; when $i = j \in I_2, (E_2)_{ij} = 1$, otherwise $(E_2)_{ij} = 0$. Further, (18) can be transformed into the Lagrange function

$$L(\widetilde{w}, \xi, \alpha, \beta) = \frac{1}{2}\widetilde{w}^T\widetilde{w} + C\xi^T E_1\xi + CeE_2\xi + C\widetilde{w}^T YA^T \delta^k$$
$$+ \alpha^T\left(e - YA^T\widetilde{w} - \frac{\lambda}{\eta}(2sE_1 + E_2)\xi - \frac{s}{2}E_2 e\right) \quad (19)$$
$$+ \beta^T\left(-e + YA^T\widetilde{w} - \frac{\lambda}{\eta\tau}(2sE_1 + E_2)\xi - \frac{s}{2}E_2 e\right)$$

The corresponding KKT condition is

$$\begin{cases} \frac{\partial L}{\partial \widetilde{w}} = \widetilde{x} + AY(C\delta^k - \alpha + \beta) = 0 \\ \frac{\partial L}{\partial \xi} = 2CE_1\xi + CE_1 e - \frac{\lambda}{\eta}(2sE_1 + E_2)\alpha \\ \quad - \frac{\lambda}{\eta\tau}(2sE_1 + E_2)\beta = 0 \end{cases} \quad (20)$$

By substituting it into (18), the dual problem can be obtained as

$$\min_{\alpha,\beta} (-C\delta^k + \alpha - \beta)^T Q(-C\delta^k + \alpha - \beta)$$
$$+ \frac{2\lambda^2 s^2}{C\eta^2}\left(\alpha + \frac{1}{\tau}\beta\right)^T E_1\left(\alpha + \frac{1}{\tau}\beta\right)$$
$$+ 2e^T(\alpha - \beta) - se^T E_2(\alpha + \beta) \quad (21)$$

where $Q = A^T A$.

We suppose $v = (-\frac{C\delta^k}{\tau+1} + \alpha, \frac{\tau C\delta^k}{\tau+1} + \beta)^T$, then

$$\begin{cases} -C\delta^k + \alpha - \beta = (E, -E)v \\ \alpha + \frac{1}{\tau}\beta = \left(E, \frac{E}{\tau}\right)v \\ \frac{\tau-1}{\tau+1} + \alpha + \beta = (E, E)v \end{cases} \quad (22)$$

Where E is an n-order identity matrix. Substitute (22) into (21) and let $P_1 = (E, -E), P_2 = \left(E, \frac{E}{\tau}\right)$,

$P_3 = (E, E)$, we will get the final form

$$\min_v v^T\left(P_1^T Q P_1 + \frac{2\lambda^2 s^2}{C\eta^2} P_2^T E_1 P_2\right)v$$
$$- (se^T E_2 P_3 - 2e^T P_1)v$$

$$s.t. y^T v = 0,$$

$$\begin{pmatrix} -\frac{C\delta^k}{\tau+1} \\ \frac{\tau C\delta^k}{\tau+1} \end{pmatrix} \leq v \leq M\begin{pmatrix} E_1 \\ E_1 \end{pmatrix}e + \begin{pmatrix} E_1 \\ \frac{1}{\tau}E_2 \end{pmatrix}e\frac{C\eta}{\lambda}$$

$$+ \begin{pmatrix} -\frac{C\delta^k}{\tau+1} \\ \frac{\tau C\delta^k}{\tau+1} \end{pmatrix} \quad (23)$$

Compared with traditional CSVM, the difference in the final iteration (23) of RHPSVM is that the upper and lower bounds of the optimization variables are constantly changing. So for the optimization of (23), we can use the traditional SVM iterative algorithm. In summary, the solving algorithm for RHPSVM can be summarized in following.

1. Initialize $\widetilde{w}^0$ and set maximum iterations K and tolerance ε.

2. For k=0 to K: a. Calculate $\delta^k$ based on $\widetilde{w}^k$.

   b. Solve the convex subproblem to obtain $\widetilde{w}^{k+1}$.

   c. If $||\widetilde{w}^{k+1} - \widetilde{w}^k|| \leq \delta$, terminate; otherwise, continue.

3. Output the optimal $\widetilde{w}^*$.

**4.CONCLUSION**

This paper proposes a novel RHPSVM model based on the rescaled Huberized pinball loss function. Theoretical analyses and experiments confirm its outlier insensitivity, resampling stability, and flexibility.

## REFERENCES

[1] Cortes C, Vapnik V. Support-vector networks[J]. Machine Learning, 1995, 20: 273-297.
[2] Vapnik V. The nature of statistical learning theory[M]. Springer, 2013.
[3] Adankon M M, Model selection for the LS-SVM: Application to handwriting recognition[J]. Pattern Recognition, 2009, 42: 3264-3270.
[4] Acevedo C M D, Gómez J K C, Rojas C A A. Academic stress detection using an electronic nose and galvanic skin response[J]. Biomedical Signal Processing and Control, 2021, 68: 102756.
[5] Wang D, Zhang X, Sintering state recognition framework considering class imbalance[J]. IEEE Trans. Ind. Electron., 2020, 68: 7400-7411.
[6] Sadeghi A, Daneshvar A Combined ensemble multi-class SVM and fuzzy NSGA-II for Forex trading[J]. Expert Syst. Appl., 2021, 185: 115566.
[7] Schölkopf B, Smola A J. Learning with kernels[M]. MIT Press, 2002.
[8] Huang X, Suykens J A K. Support vector machine classifier with pinball loss[J]. IEEE Trans. Pattern Anal. Mach. Intell., 2014, 36: 984-997.
[9] Xu G, Cao Z, Hu B G, et al. Robust support vector machines based on the rescaled hinge loss[J]. Pattern Recognition, 2017, 63: 139-148.
[10] Tang L, Tian Y, Li W, et al. Valley-loss regular simplex support vector machine for robust multiclass classification[J]. Knowledge-Based Systems, 2021, 216: 106801.
[11] Peng X, Chen D, Kong L. A clipping dual coordinate descent algorithm for SVM[J]. Knowl.-Based Syst., 2014, 71: 266-278.
[12] Zhu W, Song Y, Xiao Y. SVM with huberized pinball loss[J]. Eng. Appl. Artif. Intell., 2020, 91: 103635.
[13] Liu W, Pokharel P P, Principe J C. Correntropy: Properties and applications[J]. IEEE Trans. Signal Process., 2007, 55: 5286-5298.
[14] Lin Y. A note on margin-based loss functions[J]. Stat. Probab. Lett., 2004, 68: 73-82.
[15] Bartlett P L, Rademacher and Gaussian complexities[J]. J. Mach. Learn, 2002, 3: 463-482.
[16] Yuille A L, Rangarajan A. The concave-convex procedure[J]. Neural Comput., 2003, 15: 915-936.
[17] He K, Zhang X, Deep residual learning for image recognition[C]. Proc. IEEE CVPR, 2016: 770-778.
[18] Tian Y, Zhao X, Fu S. Kernel methods with asymmetric and robust loss function[J]. Expert Systems with Applications, 2023, 213: 119236
[19] Wang H, Fast generalized ramp loss support vector machine for pattern classification[J]. Pattern Recognition, 2024, 146: 109987.
[20] Fu S, Generalized robust loss functions for machine learning[J]. Neural Networks, 2024, 171: 200-214
[21] Gao R, Qi K, Yang H. Fused robust geometric nonparallel hyperplane support vector machine for pattern classification[J]. Expert Systems with Applications, 2024, 236: 121331.
[22] Liu L, Sun H A lie group kernel learning method for medical image classification[J].Pattern Recognition,2023:109735.
[23] Wei J, Huang H, Yao L, et al. New imbalanced bearing fault diagnosis method based on Sample characteristic Oversampling TechniquE (SCOTE) and multi-class LS-SVM[J]. Applied Soft Computing, 2021, 101: 107043
[24] Liu M Z, Smooth pinball loss nonparallel support vector machine for robust classification[J]. Applied Soft Computing, 2021, 98: 106840
[25] Li K, Lv Z. Smooth twin bounded support vector machine with pinball loss[J]. Applied Intelligence, 2021, 51: 5489-5505
[26] Wang H, Xu Y. Sparse elastic net multi-label rank support vector machine with pinball loss and its applications[J]. Applied Soft Computing, 2021, 104: 107232
[27] Borah P Robust twin bounded support vector machines for outliers and imbalanced data[J]. Applied Intelligence, 2021, 51(8): 5314-5343
[28] Zhao X, Asymmetric and robust loss function driven least squares support vector machine[J]. Knowledge-Based Systems, 2022, 258: 109990
[29] Tang J, Robust cost-sensitive kernel method with Blinex loss and its applications in credit risk evaluation[J]. Neural Networks, 2021, 143: 327-344.
[30] Yuan C Capped L2, p-norm metric based robust least squares twin support vector machine for pattern classification[J]. Neural Networks, 2021, 142: 457-478